\pdfoutput=1

\documentclass[11pt]{article}

\usepackage[final]{acl}

\usepackage{listings}

\usepackage{times}
\usepackage{latexsym}

\usepackage[T1]{fontenc}

\usepackage[utf8]{inputenc}

\usepackage{microtype}

\usepackage{inconsolata}

\usepackage{graphicx}

\graphicspath{{../figures/}}

\usepackage[acronym]{glossaries}
\makeglossaries
\newacronym{qa}{QA}{Question Answering}
\newacronym{ir}{IR}{Information Retrieval}
\newacronym{llm}{LLM}{Large Language Model}
\newacronym{rag}{RAG}{Retrieval Augmented Generation}
\newacronym{pse}{PSE}{Programmable Search Engine}
\newacronym{pmbm}{PMBM}{PubMed Best Match}
\newacronym{moe}{MoE}{Mixture of Experts}

\title{Advancing LLM detection in the ALTA 2024 Shared Task: 

Techniques and Analysis}

\author{
  \textbf{Dima Galat\textsuperscript{1}},
\\
  \textsuperscript{1}University of Technology Sydney (UTS), Australia,\\
  \small{
    \textbf{Correspondence:} \href{mailto:email@domain}{dima.galat [@] student.uts.edu.au}
  }
}

\begin{document}
\maketitle
\begin{abstract}
The recent proliferation of AI-generated content has prompted significant interest in developing reliable detection methods. This study explores techniques for identifying AI-generated text through sentence-level evaluation within hybrid articles. Our findings indicate that ChatGPT-3.5 Turbo exhibits distinct, repetitive probability patterns that enable consistent in-domain detection. Empirical tests show that minor textual modifications, such as rewording, have minimal impact on detection accuracy. These results provide valuable insights for advancing AI detection methodologies, offering a pathway toward robust solutions to address the complexities of synthetic text identification.
\end{abstract}

\section{Introduction}

The evolution of writing assistants has progressed from simple spell checkers to AI-driven systems \citep{Heidorn2000}. Advancements of \glspl{llm}, now capable of drafting entire documents, are transforming writing assistants into interactive tools capable of enhancing creativity and productivity \citep{Brown2020}. The introduction of ChatGPT has propelled \glspl{llm} into mainstream, quickly gaining them a status of a disruptive technology in many knowledge industries \citep{OpenAI2022}. ChatGPT classifier has been discontinued in 2023 seven months after launch citing low accuracy \citep{openai_classifier}.

The analysis of sentiments from early ChatGPT adopters reveals predominantly positive reactions across various domains. However, at the same time concerns have been raised regarding potential misuse and adverse effects in the context of educational activities and news media \citep{haque2022disruptive}. Being able to distinguish between human and machine-generated text is critical for maintaining integrity and transparency in academia, as well as other fields such as journalism. 

The rise of human-AI collaborative writing necessitates more advanced detection methods for analysing hybrid texts that incorporate both AI and human-authored sentences. This paper looks at the ALTA 2024 Shared Task challenge \citep{molla2024alta}, where participants  develop an automatic detection system to classify sentences in hybrid articles as either human-written or machine-generated. This paper shows ways to improve existing detection methods and promotes more responsible practices in content generation.

\section{Background and Related Work}

The strategies for a sentence-level detection task predominately focus on the following two approaches: sentence classification, where each sentence in a document is considered independently; or sequence classification, where a document is evaluated as a whole to decide labels for each word and then determine the most frequently-occurring label for the document \citep{wang2023seqxgpt}. \citeauthor{wang2023seqxgpt} proposed using token-probabilities from different \glspl{llm}, aligning local word-wise features to address differences in tokenisation, and then applying convolutions and a linear level for training a sequence classification model exhibiting strong results.

\citet{shi2024detecting} proposed to look for a boundary between AI and human-authored text, detecting transitions by modeling distances between subsequent sentences in a hybrid document. Experiments demonstrated that  this approach consistently improved classification. However, the optimal number of subsequent sentences to be evaluated depends on the document length, and additional considerations are required to account for boundaries that might exist within a hybrid sentence.

\citet{zeng2024detecting} investigated segmentation within hybrid texts to classify authorship of each segment. The findings suggest to employ a text segmentation strategy when only a few boundaries exist. Authors note that this is a challenging task; and that short texts provide limited stylistic clues, segment detection is difficult with frequent authorship changes, and human writers are free to select and edit sentences based on their preference \citep{zeng2024detecting}.

\section{Research Methodology}

Our goal is to classify sentences generated by ChatGPT-3.5 Turbo mixed with sentences written by humans. We know that at each generation step \gls{llm} predicts the next most likely token given the preceding sequence of tokens (i.e. context), or $P(\text{token}_i \mid \text{context})$. We believe that these marginal probabilities can be used to identify distinct statistical patterns in probability distributions. For example, some high-probability tokens might be favoured by an \gls{llm}, whereas human-written text would have higher entropy due to an unexpected choice of words.

%
%
%
%


%
%
%

\subsection{Data and baseline}
We were provided a training dataset of 14576 academic and 1500 news articles, containing multiple sentences with a corresponding human/machine label. Validation and test datasets contained 500 and 1000 news articles respectively. When analysing news domain data, we observed that human-written and machine-generated sentences tend to appear in continuous blocks rather than being interwoven or interspersed at the sentence level. Sequences of sentences from each class do not appear to be completely random and follow a pattern resembling the hybrid article generation method of using an \gls{llm} with \textit{fill-in task prompts} described by \citeauthor{shi2024detecting}.

 To rigorously evaluate our sentence classification capabilities without relying on contextual cues, we focus on sentence-level classification. Although a model leveraging entire article context might yield higher accuracy, such approach lies beyond the scope of our current research.

In order to evaluate the importance of domain for building a predictor we have trained baseline models using 3 different versions of the dataset:
\begin{enumerate}
\item using all of the training data 
\item using random under-sampling of academic articles to match the number of news articles
\item using only news articles	
\end{enumerate}
Our baseline model is built using a Naive Bayes classifier, chosen for the efficiency and simplicity. Best results were obtained when using TF-IDF n-gram features up to the length of 5, without stopword filtering \citep{mccallum1998comparison, ramos2003using, manning2008introduction}.  Results suggest that certain phrases and expressions can be favoured by ChatGPT. A relaxed feature independence assumption adds bias and limits the accuracy of the predictions, but this classifier is perfectly suited for comparing  statistical properties texts.

We run the model 100 times using different seed selections to account for the random under-sampling of academic articles, and for the variations of a validation data split. The results in Table~\ref{table:metrics} suggest that that using only news data is sufficient for building a sentence level predictor for this challenge. Moreover, we can see that there are distinctive statistical patterns that can be used to classify these texts.

\begin{table}[h!]
\begin{tabular*}{7.3cm}{|c|c|c|}
\hline
\textbf{Dataset}       & \textbf{Kappa Score}     & \textbf{F1 Weighted}           \\ \hline
All Data      & 0.644 $\pm$ 0.028               & 0.83 $\pm$ 0.014              \\ \hline
Sampling & 0.703 $\pm$ 0.026               & 0.86 $\pm$ 0.013              \\ \hline
Only News     & \textbf{0.716 $\pm$ 0.03}              &  \textbf{0.87 $\pm$ 0.014}             \\ \hline
\end{tabular*}
\caption{Performance metrics of Naive Bayes for different datasets based on 100 random seed selections, evaluating Cohen's Kappa Score and F1 Weighted (mean $\pm$ standard deviation).}
\label{table:metrics}
\end{table}

\subsection{Model and training}

We have attempted using \gls{llm} classification zero-shot, however this approach was not getting close to the baseline model. In order to build the best classifier we can, we selected the best base model we could adapt. 

LLaMA 3.1 \citep{meta2024llama3}, demonstrates a strong capability to generalise across various applications in natural language processing and is very popular in the research community because the release of its weights has facilitated accessibility and further experimentation. Instruction tuning, has emerged as a fine-tuning strategy which augments input-output examples with instructions, enabling instruction-tuned models to generalise more easily to new tasks \cite{wei2022}.

We are using a model variant with 8 billion parameters which can be trained on a single GPU in a few hours by using a memory-efficient QLORA \citep{qlora} training approach and 4-bit quantized weights from Unsloth \footnote{\url{https://github.com/unslothai/unsloth}}. Our best results are obtained when training in batches of 16 for 3 epochs. This model achieves 0.94 Kappa Score and 0.974 weighted F1 on our validation set, which is significantly above the baseline model results.

\section{Results}

Overall, it would appear that a 4-bit quantized LLaMA 3.1-8B-Instruct fine-tuned on a domain-specific data can be used to recognise GPT-3.5 Turbo generated content reliably based on the sentence-level evaluation alone. Table~\ref{table:kappa_accuracy} shows that our system did well in the competition, out-performing other solutions. 

\begin{table}[h]
    \centering
    \caption{Kappa and Accuracy Scores on the test set reported for the participant systems}
    \begin{tabular}{|l|l|l|}
        \hline
        \textbf{User}       & \textbf{Kappa} & \textbf{Accuracy} \\ \hline
        \textbf{our system}          & \textbf{0.9320}     & \textbf{0.9679}         \\ \hline
        samanjoy2           & 0.9080     & 0.9573         \\ \hline
        lizhuang144         & 0.8336      & 0.9235         \\ \hline
        Qihua               & 0.7605     & 0.8914        \\ \hline
        lewis\_math          & 0.6932    & 0.8565        \\ \hline
        dmollaaliod        & 0.5629     & 0.7955         \\ \hline
    \end{tabular}
    \label{table:kappa_accuracy}
\end{table}

\section{Discussion}
  
 We are still left wondering if our model can reliably detect AI-generated content to prevent misuse. Given the rapid pace of evolution of \glspl{llm}, as well as popularisation of generation strategies involving making multiple calls to \glspl{llm}, we wonder if our model would still be able to identify ChatGPT-generated sentences if we instructed another model to re-write them.

We are using the same base LLaMA 3.1-8B-Instruct we have fine-tuned for classification to re-write AI-generated sentences in our validation set. Our goal is to see if it will change our sentence classification results. We have used the following prompt, and tried running it up to two times in combination with setting a \textit{temperature} generation parameter to 0.9 to encourage randomness.

\vspace{0.5em}
\fbox{
\parbox{0.85\linewidth}{
Make re-writes to the following sentence without changing the meaning. Only return the sentence, no other information of any kind:\textbackslash n
}
}

\vspace{0.5em}
Across all of these experiments we obtained good classification results, where the lowest Kappa score produced was 0.89, and weighted F1 was 0.95. This suggests that classification is likely influenced more by the order of certain tokens than by the presence of some specific individual words.

%
%
%
  
 \section{Conclusion}
 
 
 Detecting AI-generated content is critical for maintaining authenticity and trust in written communication. We have found that given a small domain-specific corpus a fine-tuned model can reliably identify if a sentence in that corpus has been produced by GPT-3.5 Turbo. Future work could explore how this approach generalises out-of-domain, and to other \glspl{llm}.

Different models can end up producing different stylistic features, which means that some day multiple iterations of AI-edits would make it impossible to reliably judge if the text was written by a machine. For now, we have observed that AI-based sentence paraphrasing alone is inadequate to circumvent a classifier trained on in-domain samples. This highlights the importance of efforts involved in developing datasets that accurately represent the behaviour of closed-source models.

\bibliography{main}

\appendix
\onecolumn

%
%
%

\end{document}